\documentclass[final,12pt, vlined, letter, linesnumbered, ruled, boxed]{clear2023} 

\title[Meta-Causal Discovery]{A Meta-Reinforcement Learning Algorithm for Causal Discovery}
\usepackage{times}
\usepackage{booktabs}
\usepackage{bbm}

\clearauthor{%
 \Name{Andreas Sauter} \Email{a.sauter@vu.nl}\\
 \addr Vrije Universiteit Amsterdam
 \AND
 \Name{Erman Acar} \Email{erman.acar@uva.nl}\\
 \addr Universiteit van Amsterdam 
 \AND
 \Name{Vincent François-Lavet} \Email{vincent.francoislavet@vu.nl}\\
 \addr Vrije Universiteit Amsterdam%
}

\begin{document}

\maketitle

\begin{abstract}%
Uncovering the underlying causal structure of a phenomenon, domain or environment is of great scientific interest, not least because of the inferences that can be derived from such structures. Unfortunately though, given an environment, identifying its causal structure poses significant challenges. Amongst those are the need for costly interventions and the size of the space of possible structures that has to be searched. In this work, we propose a meta-reinforcement learning setup that addresses these challenges by \textit{learning} a causal discovery algorithm, called \emph{Meta-Causal Discovery}, or MCD. We model this algorithm as a policy that is trained on a set of environments with known causal structures to perform budgeted interventions. Simultaneously, the policy learns to maintain an estimate of the environment's causal structure. The learned policy can then be used as a causal discovery algorithm to estimate the structure of environments in a matter of milliseconds. At test time, our algorithm performs well even in environments that induce previously unseen causal structures. We empirically show that MCD estimates good graphs compared to SOTA approaches on toy environments and thus constitutes a proof-of-concept of learning causal discovery algorithms.  
\end{abstract}


\section{Context and Contribution}\label{sec:introduction}
Many scientific questions, from "Why did this apple fall on my head?" to "Does more physical activity reduce the risk of cardiovascular diseases?", aim at answering questions about causal effects. The field of causality offers a framework to formalize these questions. Although causality has been researched for decades \citep{ glymour1991causal, spirtes2000causation, pearl2018why}, it has recently gained momentum in the context of machine learning (ML) \citep{scholkopf2021toward} and, more specifically, reinforcement learning (RL). 

Causal models carry the promise to allow ML models to go beyond correlation-based inference through capabilities such as counterfactual reasoning and reasoning about actions. While the inference power of causal models is impressive, estimating causal structure from data, also called \emph{causal discovery}, poses several challenges.  One big challenge lies in the fact that some causal structures cannot be distinguished from observational data alone \citep{hauser2012characterization}. This issue can be mitigated by assigning values to variables independently from their causes \citep{pearl1993interv, hauser2012characterization,  bareinboim2022pearl}, a process called \emph{intervention}. Unfortunately, when confronted with real-world environments, performing interventions such as randomized controlled trials can be resource expensive or even impossible. Therefore, a large body of research exists on intervention design, or put it differently, on how to minimize the number of interventions needed to estimate the causal model.

With the successful application of RL algorithms to many domains \citep{franccois2018introduction, plaat2021high,moerland2023model}, the opportunity to use RL as a tool for causal discovery has opened as well. RL methods allow for sampling the environment as opposed to learning from a static data set of pre-collected observations. This interactive learning setting for data collection allows for estimating causal structures step-by-step considering always the newest data sample. This can be beneficial for online decisions e.g. on which variable to intervene based on how informative an intervention is for estimating a causal structure. Furthermore, an RL setting allows us to sample data from an environment, without being restricted to a fixed set of samples. This allows for better exploration of the data from an environment. 

In this work, we show that it is possible to learn an algorithm (hence the term \emph{meta-learning}) for causal discovery, called Meta Causal Discovery (MCD). More specifically, we sketch a meta-reinforcement learning model that estimates the causal structure of an environment with a given set of variables. The model is allowed to perform interventions with a limited budget to aid this process. The model simultaneously learns to perform informative interventions and to infer the updates to the structural model based on the resulting observations. During test time, the weights of the model are frozen; therefore, the model learns the causal structure only by utilizing its current network activations. Our work contributes to common challenges in causal discovery through the following capabilities:
\begin{itemize}
    \item Providing good estimates of the ground-truth causal structure compared to the SOTA.
    \item Performing causal discovery in a matter of milliseconds.
    \item Integrating observational and interventional data for causal discovery.
    \item Limiting the number of interventions through a single hyper-parameter.
\end{itemize}

\begin{figure}[h]
    \centering
    \includegraphics[width=0.75\textwidth]{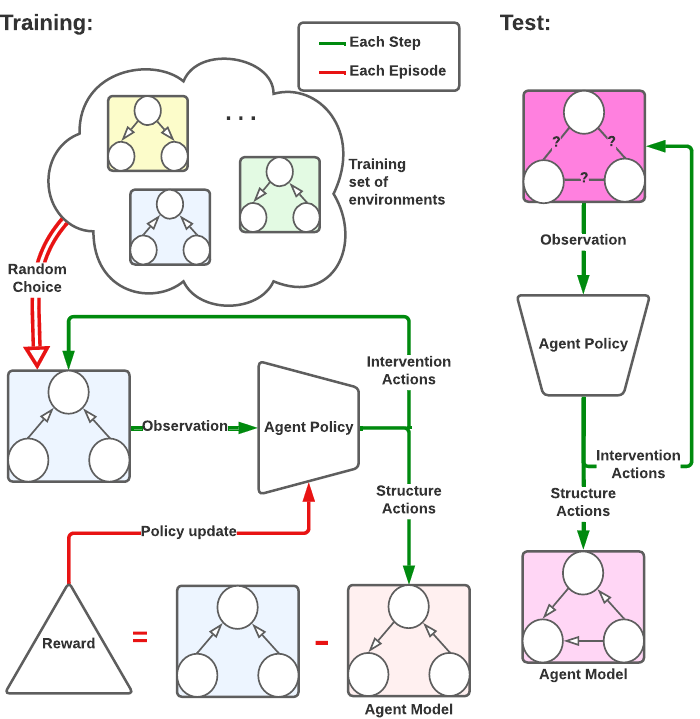}
    \caption{The learning setup of our approach. Green arrows describe processes that happen at every time step. Red arrows describe processes that happen at the end of each episode. At every time step, an observation is constructed. Based on this, our policy either performs an intervention on the environment or updates its estimate of the environment's causal structure. At the end of each episode, the estimated structure is compared to the ground truth, and the negative difference is provided as a reward. Then a new environment is sampled from the training set. During test time, the learned policy can be applied to previously unseen environments.
    }
    \label{fig:overview}
\end{figure}

We will start by introducing necessary notations (Section \ref{sec:notation}). We will then proceed to provide an overview of the relevant literature on causal discovery leading to a discussion of the common challenges of the task (Section \ref{sec:related_work}). Next, we describe our approach and our model in detail (Section \ref{sec:setup}). In a toy experiment, we show that our approach can learn to use interventions to distinguish causal structures (Section \ref{sec:proof_of_concept}). We will conclude with experiments on the accuracy of our model w.r.t. SOTA approaches and a rigorous discussion thereof (Sections \ref{sec:MCD}, \ref{sec:exp_interventions}, and \ref{sec:inter_design}). Figure \ref{fig:overview} presents an overview of our approach. The implementation can be found at \url{https://github.com/sa-and/MCD}

\section{Preliminaries and Notation}\label{sec:notation}
Causal relationships can formally be expressed in terms of a \emph{structural causal model} (SCM).  We define an SCM $S$ as a tuple  $(\mathcal{X}, \mathcal{U}, \mathcal{F}, \mathcal{P})$ where  $\mathcal{X}= \{X_1, \ldots, X_{|\mathcal{X}|}\}$ is the set of  \emph{endogenous} variables;    $\mathcal{U}= \{U_1, \ldots, U_{|\mathcal{U}|}\}$ is the set of \emph{exogenous} variables;  $\mathcal{F}= \{f_1, \ldots, f_{|\mathcal{X}|}\}$ is the set of functions whose elements are defined as  \emph{structural equations} in the form of $X_i \leftarrow f_i(.)$; $\mathcal{P} = \{P_1, \ldots, P_{|\mathcal{U}|}\}$ is a set of  pairwise independent distributions where $U_i \sim P_i$. Every SCM induces a \emph{graph structure} $G$ in which each node represents a random variable. For all nodes $ W_i \in \{\mathcal{X}\cup \mathcal{U}\}, X_j \in \mathcal{X}$, the induced graph $G$ has a directed edge $(W_i, X_j)$ iff $W_i$ is an input of $f_j$. This implies that every exogenous variable $U_i$ is a root node in $G$. In this work, we restrict ourselves to SCMs that induce a \emph{directed acyclic graph} (DAG).

An intervention on a variable $X_i \in \mathcal{X}$ is defined as replacing the corresponding structural equation $X_i \leftarrow f_{i}(.)$ with $X_i \leftarrow x$ for some value $x$, which we denote as $do(X_i=x)$. Intervening on a variable makes it independent of its parents and removes its incoming edges in $G$. The model is causal in the sense that one can derive the distribution of a subset  $\mathcal{X}' \subseteq \mathcal{X}$ of variables following an intervention on a set of variables, called \emph{intervention target}, $\mathcal{I} \subseteq \mathcal{X} \setminus \mathcal{X}'$. We call the resulting distribution over $\mathcal{X}$ \emph{post-interventional}. When no intervention is performed ($\mathcal{I}=\emptyset$) we call the resulting distribution an \emph{observational} distribution.

We characterize our RL learning problem by a \emph{state-space} $\mathcal{S}$, an \emph{observation space} $\Omega$, an \emph{action-space} $\mathcal{A}$, a \emph{reward} function $r(s): \mathcal{S} \rightarrow \mathbb{R}$, and a \emph{policy} $\pi (h): \Omega^t \rightarrow \mathcal{A}$, where $h$ is the history of observations up to time step $t$. We define an \emph{episode} $e$ as the state-action sequence from the beginning to the end of the estimation. We will refer to the length of the episode as \emph{horizon} $H$. The \emph{value function} $V_\pi(h): \Omega^t \rightarrow \mathbb{R}$ defines the expected, discounted cumulative reward following a deterministic policy $\pi$, with \emph{discount factor} $\gamma$. The objective of the RL agent is to find the optimal policy $\pi^*$ that maximizes the value function for all observations which can be expressed as $\pi^*(h) = \text{argmax}_\pi V_\pi(h), \forall h \in \Omega^t$. We describe our approach as a meta-learning setup, since we use RL, to learn a policy that learns a causal structure of an environment.

\section{Related Work}\label{sec:related_work}
Due to its relevance in many applications, causal discovery research has gained momentum in the last years leading to an impressive body of work \citep{vowels2021d}. Score-based causal discovery approaches search the space of DAGs via metrics that indicate how well the graph fits the data. This is often done greedily over the space of classes of graphs in which the graphs can only be distinguished via interventions \citep{meek1997graphical,chickering2002optimal,  hauser2012characterization,ramsey2017million}, or over permutations of node orderings \citep{solus2017consistency, wang2017permutation,yang2018characterizing}. Constraint-based approaches leverage the statistical independence patterns in the data to constrain the possible output graphs \citep{ glymour1991causal,spirtes2000causation}. These constraints can even be expressed as propositional formulas and then solved with answer-set programming \citep{hyttinen2014constraint}. RL offers an alternative way of searching the space of DAGs by using the reward to navigate toward good graph generators \citep{zhu2019causal}. Note that many algorithms rely on strong assumptions on the class of causal relations e.g. linear additive noise models \citep{buhlmann2014cam, peters2014causal, shimizu2006linear}. This makes these algorithms interesting for theoretical analysis but it also restricts their application potential in practice. 

Since the number of possible DAGs grows super-exponentially in the number of nodes \citep{robinson1977counting}, most  score- and constraint-based approaches suffer from long run times. A recent line of research tackles this problem by deploying optimization-based algorithms. These algorithms work e.g. with constraint optimization \citep{zheng2018dags, brouillard2020differentiable} but also by learning causal graph neural networks \citep{goudet2018learning, yu2019dag, ton2021meta} or variational auto-encoders \citep{Yang_2021_CVPR}. For neuro-causal models, advances are also made in the theoretical analysis of their identifiability \citep{xia2021the}. A similar approach is taken by works that sample both the graph structures and the functional parameters from posterior distributions \citep{ke2019learning, lippe2021efficient, scherrer2021learning, lei2022causal}. This improves learning efficiency, not only of the structures but also of the functional relations of the causal mechanisms. While optimization-based approaches can reduce the run-time for structure learning by avoiding a combinatorial explosion, they can still take significant time to learn the causal structure. In our work, we are shifting the computational complexity to training time to circumvent long runtimes at test time.

Another common challenge amongst most causal discovery algorithms is the integration of observational and interventional data. Although integrating frameworks exist \citep{mooij2020joint}, only a fraction of causal discovery algorithms successfully jointly consider interventional and observational data \citep{vowels2021d}. A promising direction for the seamless integration of interventional data is by means of RL. We argue that this is partly due to the implicit connections between interventions and actions in any RL framework, and partly because RL can easily be combined with deep-learning models. Our work distinguishes itself from these closely related works in different ways. While \cite{dasgupta2019causal} developed an algorithm that is similar to ours, their primary task was not causal discovery. \cite{nair2019causal}, \cite{gasse2021causal}, \cite{lei2022causal}, and \cite{mendez2022causal} put a strong focus on using causal structures to aid RL while learning the structures is done in a supervised manner.  Similarly, \cite{scherrer2021learning} and \cite{tigas2022interventions} develop an active learning algorithm that chooses interventions more efficiently to estimate the structure from this data. \cite{amirinezhad2022active} have a similar setup and task but restrict RL to learn a heuristic function for choosing the next intervention target. Furthermore, they do not take into account the values and distributions of the random variables. Their graph-updating procedure is pre-defined, whereas in our approach the update rules are learned.

\section{Meta-Reinforcement Learning Setup}\label{sec:setup}
\subsection{Actions}\label{sec:actions}
We implement two types of discrete actions. The first type performs an intervention on the current environment SCM. This enables the policy to choose a (post-interventional) distribution to sample from.  We will refer to this kind of action as \emph{listening action}. All, except for one, of the listening actions are \emph{intervention actions} that intervene on exactly one variable (i.e., $|\mathcal{I}|=1$). For each endogenous variable $X \in \mathcal{X}$, we provide an action $do(X=c)$ for a constant $c$. We argue that $c$ should be chosen in a way that makes it easy to distinguish the post-interventional distribution from the observational distribution i.e. it should be unlikely that samples from the post-interventional distribution come from the observational distribution. A future expansion of our work could include learning a good $c$. The intervention actions amount to a total of $n$ actions for $n$ nodes. There is one additional listening action which we call the \emph{non-action}. When the non-action is taken, the agent does not intervene (i.e., $\mathcal{I}=\emptyset$). This action accounts for the collection of purely observational data.

The second type of action is responsible for maintaining the current causal structure estimate of the environment, which we call the \emph{epistemic model}. We will refer to these actions as \emph{structure-actions}. Each structure action can either \emph{add}, \emph{delete} or \emph{reverse} an edge of the epistemic model. Whenever a delete or reverse action is applied to an edge that is not present in the current model, the action is ignored. This is effectively equivalent to performing the non-action. The same holds when the add action is applied to an edge that is already in the epistemic model. We do not make any further restrictions, for instance, w.r.t. acyclicity for the structure actions. 

For a graph with $n$ nodes, there are $n(n-1)$ possible edges, and hence there are $3n(n-1)$ structure actions. Together with the listening actions we have $n + 1 + 3n(n-1)$ actions. Therefore, the size of the action space is quadratic in the size of nodes.  

\subsection{Observation Space}\label{sec:state_space}
In this Section, we describe how observations $o \in \Omega$ are constructed. Each environment is completely defined by an SCM. At each time step $t$, the exogenous variables $\mathcal{U}$ are sampled. The functions $\mathcal{F}$ are then evaluated according to the topological ordering of their corresponding nodes in $G$. This results in a sample of the endogenous variables $\mathcal{X}$ and makes up the first part $o^V_t$ of the observation vector.

The second part of the observation, $o^A_t$ is a one-hot vector that indicates the intervention target. If the $i$-th element of $o^A_t$ is $1$, then there is an intervention on $X_i$. 

The third part of the observation, $o^G_t$ encodes the current epistemic model as a vector. Each value of this vector represents an undirected edge in the graph. The edges in the vector are ordered lexicographically. The value 0 encodes that there is no edge between the two nodes. The value  0.5 encodes that there is an edge going from the lexicographically smaller node to the bigger node of the undirected edge. And the value 1 encodes that there is an edge in the opposite direction. For example, a 3-node graph $X_0 \rightarrow X_2 \rightarrow X_1$ would be encoded as $o^G_t = [0, 0.5, 1]$. 

The last element in the observation, $o^T_t$, encodes the time until the end of an episode normalized to 1 as $o^T_t=\frac{t}{H}$, where $H$ is the horizon. The complete observation $o_t$, is the concatenation of $o^V_t, o^A_t, o^G_t,$and $ o^T_t$ as shown in Figure \ref{fig:agent_policy}. Taken together, the size of one observation is $2n + n(n-1)/2 + 1$ with $n$ endogenous variables and hence quadratic in the size of the graph. The input to our policy is the history of observations from the beginning of the episode $h_t=o_{0:t}$ which we will approximate by implementing a recurrent policy (see Section \ref{sec:algo_and_policy}).

\subsection{Rewards and Episodes}\label{sec:rew_eps}
Our task is to find the causal structure of the environment, i.e., the DAG that corresponds to the graph induced by the SCM of the environment. Therefore, we compare the epistemic model to the true causal structure of the environment. The quantification of this comparison serves as the reward for our algorithm. We count the edge differences between the two graphs. This ensures that generating a model that has more edges in common with the true DAG will be preferred over one which has fewer edges in common. It further gives a strong focus on causal discovery as opposed to scores based on causal inference. Specifically, we use a variant of the \emph{Structural Hamming Distance} (SHD) \citep{tsamardinos2006max}. In this variant, we take two directed graphs and count how many of the edges need to be removed or added to transform the first graph into the second graph. This results in a metric that simply counts the distinguishing edges of two directed graphs. We will refer to this metric as \emph{directed SHD} or \emph{dSHD}. Given a predicted directed graph $G_P = (V, E_P)$ and a target, directed graph $G_T = (V, E_T)$, we define the dSHD as $dSHD(E_P, E_T) = \mid E_P \setminus E_T \mid + \mid E_T \setminus E_P \mid$.

For each episode, we set a finite horizon $H$. The estimation of the epistemic model is complete when $H-1$ actions were taken. Dynamically determining the end of the estimations is left for future research. Note that when a small episode length is chosen, fewer samples can be collected by the agent. This might impact how well the agent is informed on which updates to make to the epistemic model. At the same time, $H$ should not be set too large since additional learning complexity might be introduced. At the beginning of each episode, an SCM is sampled from the training set and the epistemic model of the agent is reset to a random DAG, to further introduce randomness. 

The reward is calculated by taking the negative dSHD between the generated DAG and the true causal graph at the end of each episode. Every other step receives a bonus of 0.1 if an intervention action is performed. This will lead to an algorithm that performs more interventions. However, our main parameter to reduce interventions is the horizon, which puts a hard cap on the maximum number of interventions. We introduced the bonus for interventions because it worked well as a reward-shaping tool. The resulting value function for a history of observations $h$ and a policy $\pi$ is then defined as
\begin{equation}
V_\pi(h)=\mathbb E_{h \sim \pi} \left[ -\gamma^{H-t}\text{dSHD}(E_{Epi}^H, E_{Env})   \mid h_0 = h\right] + E_{h\sim \pi} \left[\gamma^t 0.1\mathbbm{1}_I(h)\mid h_0 = h\right] 
\end{equation}
where $E_{Epi}^H$ are the edges of the epistemic model at the end of an episode, $E_{Env}$ are the edges of the ground truth causal graph and $\mathbbm{1}_I(h)$ is the function that indicates whether there is an intervention in the latest observation in $h$. An optimal policy on this value function will construct an epistemic model that corresponds to the DAG induced by the causal structure of the current environment. The pseudocode of our overall learning setup can be seen in Algorithm \ref{algo:MCD}.

\begin{algorithm}[h]
    \hrulefill \BlankLine
    \KwIn{test set of DAGs $D$, number of training episodes $N$, horizon $H$}
    \KwOut{causal discovery policy $\pi$}
    \BlankLine
    \emph{initialize policy $\pi$} \tcp*{(Section \ref{sec:algo_and_policy})}

    \For{$ep \leftarrow 0; ep<N; ep \leftarrow ep + 1$}{
        \emph{generate random SCM $S$ with $G \notin D$} \tcp*{New training env  (Section \ref{sec:rew_eps})}

        \emph{initialize $h$ and a random epistemic model $\hat{G}$} \tcp*{(Section \ref{sec:state_space} and \ref{sec:rew_eps})}
        \BlankLine
        \For{$steps \leftarrow 0; steps<H; steps\leftarrow steps + 1$}{  
            \emph{$r \leftarrow 0$} \tcp*{Reset reward}
            
            \emph{$a \leftarrow \pi(h)$} \tcp*{Determine action based on policy}
            
            \BlankLine
            \eIf{\emph{$a$ is structure action}}{
                \emph{$\hat{G}$.update($a$)} \tcp*{Update epistemic model (Section \ref{sec:actions})}
            }
            {\emph{$S$.do($a$)} \tcp*{Intervene on SCM $S$  (Section \ref{sec:actions})}

            \emph{$r \leftarrow r + 0.1$} \tcp*{Bonus for interventions (Section \ref{sec:rew_eps})}
            }
            
            \emph{$h.update([S.sample() + a.onehot() + \hat{G}.encode() + [\frac{steps}{H}]]$)} \tcp*{Update observation history (Section \ref{sec:state_space})}

            \BlankLine
            \If{$steps == H-1$}{
                 \emph{$r \leftarrow r - dSHD(\hat{G}, G)$} \tcp*{Compare $\hat{G}$ to true $G$ (Section \ref{sec:rew_eps})}
            }
            \emph{$\pi.update(r)$}
        }        
    }
    \KwRet{$\pi$}

    \hrulefill
\caption{Pseudocode of our MCD learning algorithm. The returned policy $\pi$ can be applied to new environments without re-training.}
\label{algo:MCD}
\end{algorithm}

\subsection{Learning Algorithm and Policy Network}\label{sec:algo_and_policy}
We use the Actor-Critic with Experience Replay (ACER) \citep{wang2016sample} algorithm to learn our policy. We choose this algorithm because of its sample-efficient off-policy method, its (potentially) easy extension to continuous action spaces, and because it worked well after preliminary experiments. We use a discount factor $\gamma = 0.99$, a buffer size of 500000, and a constant learning rate. All other parameters are according to the standard values of Stable-Baselines \citep[version 2.10.1]{stable-baselines}.

\begin{figure}
    \centering
    \includegraphics[width=0.65\textwidth]{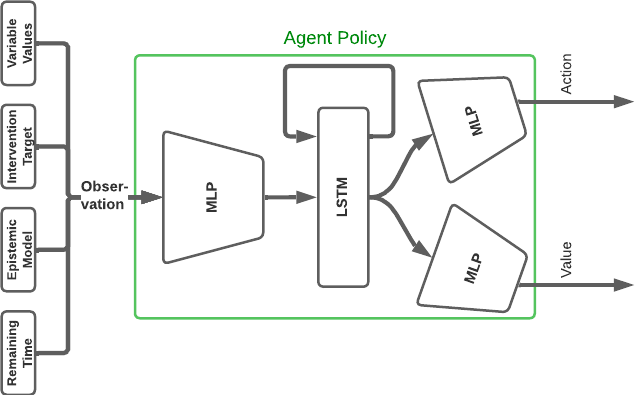}
    \caption{An overview of our observations and policy network. The observation consists of the environment's current variable values, the intervention target, the encoding of the current epistemic model, and the remaining time as a fraction of the horizon. The actor and the critic share a fully connected feed-forward network with an additional LSTM layer that approximates the history of observations.}
    \label{fig:agent_policy}
\end{figure}

The architecture of our policy network is sketched in Figure \ref{fig:agent_policy}. Both, the actor-network and the critic-network are fully-connected multi-layer perceptrons (MLP). They are preceded by a shared MLP for feature extraction and a single LSTM layer \citep{hochreiter1997long}. We introduce the LSTM layer to allow the policy to estimate the history of observations as described in Section \ref{sec:notation}. With the LSTM, our policy can access an aggregated version of previous variable-intervention pairs. This can help to disambiguate post-interventional distributions and has been shown to work in similar research \citep{dasgupta2019causal}. The exact amounts of layers and their sizes are specified for each experiment. 

\section{Learning to Intervene}\label{sec:proof_of_concept}
To test whether our approach can learn to perform the right interventions to identify causal models under optimal conditions we develop a toy example. To this end, we construct a simple setup in which two observationally equivalent, yet interventionally different environments have to be distinguished. This means that the causal structures of the two environments can only be distinguished by intervening on them \citep{bareinboim2022pearl}. Observing the values of the variables is not enough for distinguishing their structure. For this experiment, we disable the bonus reward for performing interventions. Thus, if the policy should distinguish the two environments, it has to learn that interventions are needed and that certain structures can be inferred from those interventions.

The two environments are governed by SCMs with 3 endogenous variables $X_1, X_2, X_3$, and structures $G_1: X_1 \leftarrow X_0 \rightarrow X_2$ and $G_2:  X_0 \rightarrow X_1 \rightarrow X_2$. In both environments, the root node $X_0$ follows a normal distribution with $X_0 \sim N(\mu = 0, \sigma = 0.1)$. The nodes $X_1$ and $X_2$ take the values of their parents in the corresponding graph. The resulting observational distributions $P_{G_1}(X_0, X_1, X_2)$ and $P_{G_2}(X_0, X_1, X_2)$ are equivalent and so are the post-interventional distributions after interventions on $X_0$ or $X_2$. For an intervention on $X_1$, $P_{G_1}(X_0, X_2 \mid do(X_1=x)) \neq P_{G_2}(X_0, X_2 \mid do(X_1=x))$. Hence the two SCMs can only be distinguished by intervening on $X_1$. The details for the training setup can be found in Appendix \ref{apx:setup_learning_to_intervene}. The algorithm is trained and evaluated in both environments, where at the beginning of each episode, one of the two environments is picked at random. This setup allows us to investigate whether, given enough training time and data, our approach \emph{can} learn to distinguish observationally equivalent environments. 

\begin{figure}[h]
    \centering
\includegraphics[width=0.9\textwidth]{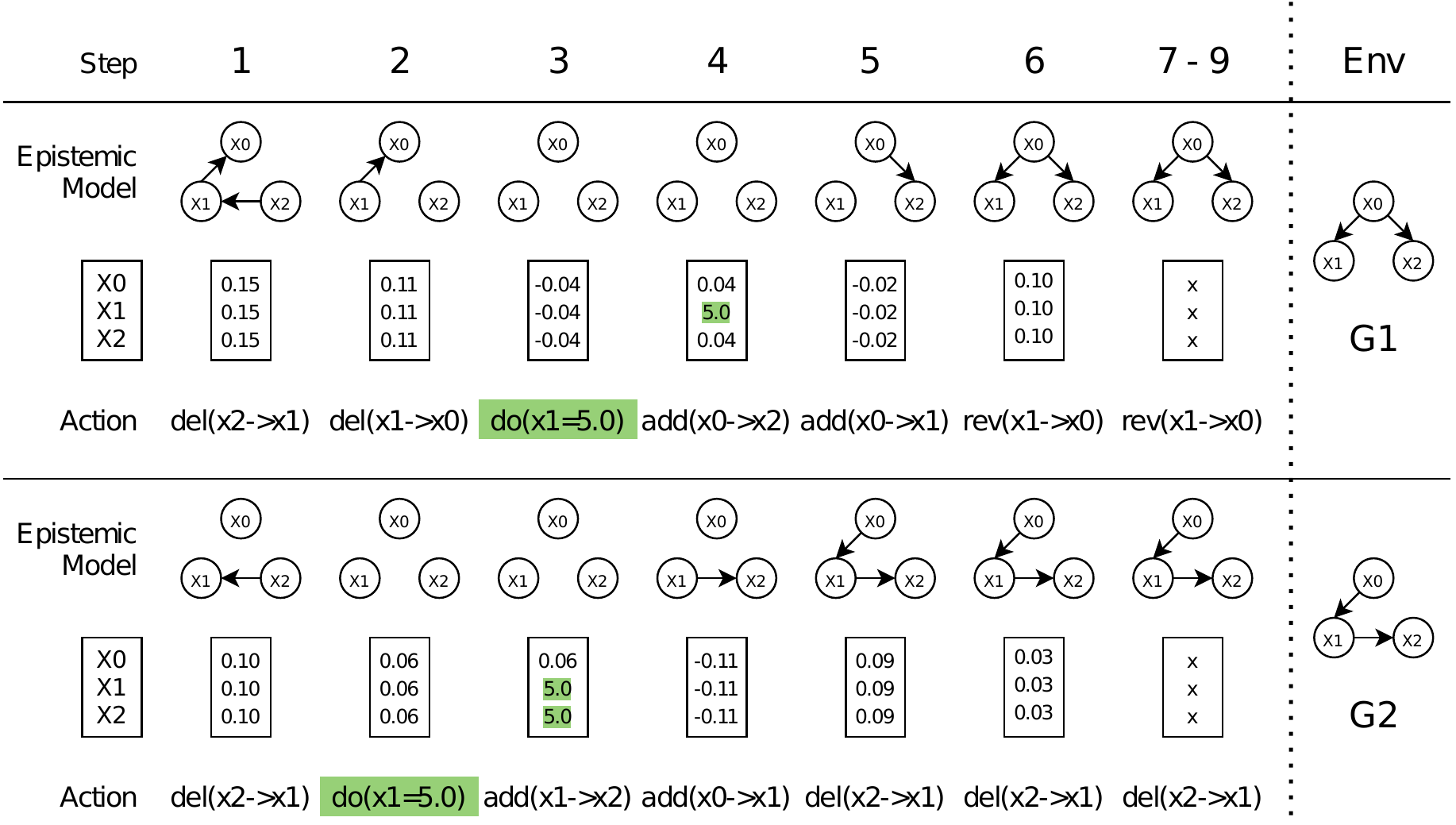}
    \caption{Illustration of two sample episodes after training with the respective causal environments $G_1$ and $G_2$. Each step shows the current epistemic model, the current values of the three random variables, and the action which is chosen by the policy based on those observations. Interventions and their effects are highlighted in green. In steps 7-9 neither the epistemic model nor the resulting action changes.}
    \label{fig:proof_of_concept}
\end{figure}

After training, we observe that the mean dSHD between the generated epistemic models and the ground truth graphs is 0.0 with a standard derivation of 0.0. This is a perfect reproduction of the two environments in all cases. This indicates that our policy has indeed learned to use the right intervention to find the true causal structure. For further testing, we apply the converged policy 10 times to each of the two environments and qualitatively analyze the behavior. What the resulting 20 episodes have in common is that, towards the beginning of each episode, they tend to delete edges that do not overlap in the two environments. Then an intervention on $X_1$ is performed. Depending on the outcome of the intervention, either $G_1$ or $G_2$ is ultimately generated. This can also be seen in the two hand-picked example episodes in Figure \ref{fig:proof_of_concept}.

These results show that our learned policy learns to use the intervention on $X_1$ to distinguish between the two environments. Thus, our approach is capable of learning to use interventions in an active manner and generating the appropriate graph from the resulting observations. If this can be successfully learned in more complex environments, our learned policy could potentially be used to discover new rules of causal structure estimation. Furthermore, these results suggest that the model has learned to only perform interventions that are relevant to causal discovery as opposed to random interventions. 

\section{Learning a Causal Discovery Algorithm}\label{sec:MCD}
In this section, we investigate whether the agent can learn a causal discovery algorithm with the meta-learning setup described in Algorithm \ref{algo:MCD}. More specifically, we train our policy on a set of SCMs where we sample a new SCM at every episode at random.  If learning such a policy is successful, the policy accurately estimates the causal structure of these environments, even when the weights are frozen and the ground truth structure is unknown. We hypothesize that the learned policy is a causal discovery algorithm in itself, and can thus successfully be applied even in environments with previously unseen causal structures. 

We test this by running the learned policy, with frozen weights, on a test set of SCMs and compare it to the  SOTA causal discovery algorithms ENCO \citep{lippe2021efficient}, DCDI \citep{brouillard2020differentiable}, NOTEARS \citep{zheng2018dags}, and a baseline that generates a random DAG. Note that ENCO and DCDI can both integrate observational and interventional data while NOTEARS only uses observational data. 

Following the widely adopted practice, we test our approach on SCMs that have an additive linear causal model with independent Gaussian noise. Although this choice limits the applicability to real-world environments, it provides a good means for comparison to other approaches. To make our approach more general, a variant could be learned in which the training SCMs have more general functional relations. It is known that environments with linear additive functions and Gaussian noise can suffer from varsortability where good results can be achieved by ordering the variables by the variance of their observational distribution \citep{reisach2021beware, kaiser2021unsuitability}. To make our approach less prone to this error, we randomly sample the variance of each Gaussian noise we use.  

Given a structure $G=(V, E)$ and $\mathcal{X}=V$, we model our SCM environments as follows. Each exogenous variable $U_i$ follows a distribution $P_i = N(\mu = 0, \sigma = \Sigma)$ where $\Sigma$ is sampled from $Uniform([0; 0.5])$ for every $U_i$. For each endogenous variable $X_i$, we model $f_i$ as
\begin{equation}\label{eq:env}
    f_i(Pa_X^G, U_i) = \left(\sum\limits_{Y \in Pa_X^G} WY \right)+ U_i
\end{equation}
where $Pa_X^G$ are the parents of $X$ in $G$, and $W \sim Uniform([-1; 1])$ represents a random weight for each causal effect of a parent to a child. 

We randomly generate a test set of 7 DAGs with 3 variables and 200 DAGs with 4 variables. We generate 10 SCMs following Equation \ref{eq:env} for each of these graphs. During training, we generate a random ground truth DAG at the beginning of each episode. If this random DAG is in the test set, we discard it and sample a new random DAG. This process is repeated until the sampled DAG is not in the test set to ensure that our model has never seen the causal structures in the test set. When a training DAG is found, we generate an SCM as described above as our current environment. We chose to generate the training set this way, so our training set covers as much of the space of DAGs as possible. The training details for our policy can be found in Appendix \ref{apx:setup_learning_MCD}. We will refer to the best model that is found during training as \emph{best model}. 

To compare our approach, we used the following setup for the benchmarks. For NOTEARS \citep{zheng2020notears} we sampled 10000 samples from the observational distribution of each SCM. For ENCO \citep{lippe2021efficient} we sampled 10000 samples from the observational distribution and 3333 samples from each post-interventional distribution (one per variable) and trained for 50 epochs. For DCDI \citep{brouillard2020differentiable} we took 3333 samples from each post-interventional distribution as well and trained the deep sigmoidal flow model version of the algorithm for 50000 iterations. We use the original implementation from the authors of the corresponding papers.  For each of the algorithms, we computed the dSHD between the predicted DAG and the ground truth DAG. Table \ref{tab:results_normal_run} shows the results of running our best model (with frozen weights) and the benchmarks on the first 50SCMs in the test set.

\begin{table}[h]\centering 
\begin{tabular}{@{}lcccccc@{}}
\toprule
           & \multicolumn{3}{c}{3 Variables} & \multicolumn{3}{c}{4 Variables} \\ \midrule
           & mean     & median     & std     & mean     & median     & std     \\
Random     & 4.43     & 4.0        & 0.90    & 4.80      & 5.0        & 1.72    \\
DCDI       & 2.94     & 3.0        & 0.70    & 4.44     & \textbf{4.0}        & 1.77    \\
ENCO       & 3.18     & 3.0        & 1.09    & 3.74     & \textbf{4.0}        & 1.73    \\
NOTEARS    & 2.50      & 3.0        & 0.92    & 3.72     & \textbf{4.0}        & 1.77    \\
\textbf{MCD (ours)} & \textbf{1.28}     & \textbf{1.0 }       & \textbf{0.66}    & \textbf{3.60}      & \textbf{4.0}        & \textbf{1.62}    \\ \bottomrule
\end{tabular}
\caption{Statistics over the dSHDs resulting from running the algorithms on the first 50 SCMs in the test set.}
\label{tab:results_normal_run}
\end{table}

Firstly, Table \ref{tab:results_normal_run} shows that our approach outperforms the random baseline, suggesting that MCD learns to estimate the environment's causal structure beyond randomly orienting edges. The means over the resulting dSHDs suggest that our approach compares favorably to the benchmarks. To investigate this difference in more detail, we performed a one-sided Wilcoxon signed-rank test between the estimates from our policy and the estimates from DCDI, ENCO, and NOTEARS. To correct for performing 3 comparisons, we consider a significance level of 1.7\% instead of 5\%. In the 3 variable case as well as the 4 variable case we can conclude that the dSHDs from our method are significantly lower (all p-values $<$ 1.7\%) than the ones from any of the other algorithms. Please note that the results for ENCO are somewhat unexpected. Refer to Appendix \ref{apx:enco_discussion} for an elaboration on the issue. 

We note that each of these estimations of MCD takes an average of 23ms in the 3 variable case and 30ms in the 4 variable case on a consumer-grade notebook. This is in contrast to the SOTA, which can take minutes for one estimation. We attribute this performance to the fact that one estimation of MCD only takes $H$ forward passes through the policy network, and that the computational complexity of our approach is shifted completely to training time.

We conclude that with our approach a causal discovery algorithm can be learned that interactively performs interventions and updates its structure estimate. Our algorithm not only compares favorably to the SOTA w.r.t. to the dSHD to the ground truth graph but is also computationally quick in deriving the estimate making it interesting for a variety of applications.

\section{Contribution of Interventions}\label{sec:exp_interventions}
To empirically investigate the effect of interventions on the performance of our algorithm, we perform an ablation study. To this end, we train a variant of our policy (MCD-O) which is based on purely observational data, i.e. we disallow the use of interventions. We then compare our results to the results of MCD and NOTEARS, which also works on purely observational data and the random baseline of the previous section. We decided to compare MCD-O to NOTEARS, to investigate if, even in the observational setting, the learned policy constitutes a good causal discovery algorithm.

We train our model with the same parameters as in Section \ref{sec:MCD} and measure the dSHD on the first 50 3-variable SCMs in the test set with the best model of the training run and frozen weights. We perform a Wilcoxon signed-rank test to evaluate whether there is a significant difference between the model that uses interventions and the one that does not. We also test whether there is a difference between NOTEARS and our approach when no interventions are allowed. 

The statistics of MCD-O applied once on the first 50 test SCMs are as follows: mean=2.6, median=3.0, std=1.44. When comparing this to the version which uses interventions (mean=1.28, median=1.0, std=0.66, see Table \ref{tab:results_normal_run}), we can see the importance that interventions have on the overall performance of MCD. This is confirmed by performing a Wilcoxon signed-rank test between the results of MCD and MCD-O indicating that MCD is significantly better (with $p << 0.025$). When comparing MCD-O with NOTEARS, we do not observe any significant difference in a two-sided Wilcoxon signed-rank test ($p \sim 0.4$). In other words, while MCD-O does not provide an improvement over NOTEARS, it still constitutes a valid alternative approach. These results confirm that introducing interventions results in the hypothesized edge over the purely observational version of our model.

\section{Aspects of Intervention Design}\label{sec:inter_design}
As argued in Section \ref{sec:introduction}, MCD provides an approach to restrict the number of interventions needed to accurately discover the causal structure of an environment. The upper bound of interventions that the learned policy will perform is the horizon of an episode (20-30 in our experiments). Compared to the interventions used in the benchmarks (up to 10000 samples from the observational distribution and up to 3333 samples from the interventional distributions), this is a significant improvement considering the comparably good performance of MCD. 

Empirically we see that our best model from section \ref{sec:MCD} performs an average of 17 interventions in the 3-variable environments. On average, 64\% of the interventions were on the first variable and 36\% on the third variable. No interventions were performed on the second variable in any of the runs. To investigate this behavior, we ran checkpoints of the model of earlier steps of the training and found that the model is performing interventions on the second variable in those checkpoints. We hypothesize that the model learns to not do this intervention because of the composition of the training SCMs.

To have a better comparison of the performance of MCD in a context where interventional samples are hard to obtain, we re-evaluate our approach w.r.t. SOTA. We run the benchmarks described in section \ref{sec:MCD} again, with a different number of samples to see how they perform under comparable sample sizes. For DCDI, we take 17 samples from each post-interventional distribution. For ENCO we take 17 samples from each post-interventional distribution and 4 samples from the observational distribution. For NOTEARS we take 20 observational samples. As before, we run the benchmarks on the first 50 SCMs in the test set of 3 variables and report the statistics of the dSHD to the ground truth graph.

\begin{table}[h]
\centering
\begin{tabular}{@{}lccc@{}}
\toprule
           & mean     & median     & std     \\ \midrule
DCDI       & 3.46     & 3.0        & 1.00    \\
ENCO       & 2.40      & 3.0        & 0.80    \\
NOTEARS    & 2.84     & 3.0        & 1.02    \\
\textbf{MCD (ours)} &\textbf{ 1.28}     & \textbf{1.0}        & \textbf{0.66}    \\ \bottomrule
\end{tabular}
\caption{Statistics over the dSHD obtained from predicting the causal structure of the 3 variable test SCMs by algorithm. The samples sizes for DCDI, ENCO, and NOTEARS are reduced to approximately match the number of samples used by MCD.}
\label{tab:results_design}
\end{table}

Table \ref{tab:results_design} shows the statistics over the dSHD obtained from running the corresponding algorithms on the 3 variable test SCMs. As expected, we see an increase in the performance gap between MCD and DCDI and MCD and NOTEARS. This indicates that their ability to perform well when few interventions are provided is limited for DCDI and NOTEARS. The better performance of MCD w.r.t. ENCO can also still be seen, although ENCO seems to perform slightly better in this low-sample regime (see Appendix \ref{apx:enco_discussion} for a discussion). 

These investigations show that SOTA causal discovery algorithms rely on many samples to improve their prediction accuracy. At the same time, MCD generates more accurate DAGs w.r.t. the SOTA, while using only a fraction of samples. Furthermore, when confronted with a sample size that is similar to MCD, DCDI, and NOTEARS quickly drop in performance. Overall, we conclude that MCD uses interventions in an efficient way which makes it perform well even when the budget for interventions is low. The low number of interventions needed for MCD promises to make it more applicable than SOTA, especially in domains in which interventions are costly. 

\section{Conclusion} 
This paper presents an approach to learning a causal discovery algorithm. In our RL setting, we learn a policy that simultaneously learns to perform informative interventions and update an estimate of the causal structure of the environment. Once the policy is learned, it can be used to perform causal discovery even on environments whose structure it has not encountered during training in a matter of milliseconds. This is partly because of its ability to integrate interventional and observational data. By limiting the episode length, we put an upper bound on the number of interventions that can be performed by MCD, making it more suitable for applications where interventions are costly.
 
We acknowledge that our approach needs modifications to scale to realistic environments with more variables. The explosion of the action- and state-space that this would imply prompts considerations about better encodings. A further problem in a potential real-world setting is the availability of a large amount of data-generating models for training. To perform well on all the possible causal relations in the real world, the class of training SCMs would need to be significantly expanded. An alternative approach would be to make MCD transferable to SCM classes other than linear-additive SCMs. We argue that also an extension to a scenario in which the variables are learned from raw input would lead to even better applicability since hand-crafted variables often introduce sub-optimalities w.r.t. task performance.

\acks{We thank Aske Plaat and the anonymous reviewers, whose suggestions helped improve the final version of this paper significantly. Furthermore, we thank Philip Lippe for taking the time to share insight on how to tune ENCO. \par This research was partially funded by the Hybrid Intelligence Center, a 10-year programme funded by the Dutch Ministry of Education, Culture and Science through the Netherlands Organisation for Scientific Research, \url{https://hybrid-intelligence-centre.nl}, grant number 024.004.022} 

\bibliography{references}

\appendix

\section{Training Details}
\subsection{Learning to Intervene}\label{apx:setup_learning_to_intervene}
For the experiment in section \ref{sec:proof_of_concept} the policy network has a fully connected layer of size 30, followed by an LSTM layer of size 30. The actor-network has one fully connected layer of size 30, and the critic-network has one fully connected layer of size 10. The length of each episode was set to 10 and the model was trained for 5 million training steps. As intervention actions we provided $do(X_i=0)$ and $do(X_i=5)$ for each $X_i \in \mathcal{X}$.  For all other parameters, the default values were used. 

\subsection{Learning a Causal Discovery Algorithm}\label{apx:setup_learning_MCD}
The following configuration for the policy network of the experiment in Section \ref{sec:MCD} worked best after preliminary experiments for the 3-variable (4-variable) environments: One (two) fully connected layer(s) of size 30 (64) followed by an LSTM layer of size 30 (128). Its outputs are fed into a fully connected layer of size 30 (32) for the actor-network and one of size 10 (32) for the critic-network. For this experiment, we set the horizon to 20. As intervention actions we provide $do(X_i=5)$ for each $X_i \in \mathcal{X}$. We chose this value since it is unlikely to come from any of the noise distributions.

\section{Results for ENCO}\label{apx:enco_discussion}
The results in Section \ref{sec:MCD} and \ref{sec:inter_design} raised some questions about the correctness of our comparison. More specifically, two anomalies for the ENCO \citep{lippe2021efficient} benchmark emerged. Firstly, in all runs, ENCO seems to perform worse than the reports in the original paper might suggest. Secondly, when decreasing the number of samples in the training set, ENCO improves in performance. Neither of these behaviors is expected or can be easily explained.

We investigated these issues in more detail and contacted one of the authors of the original paper for a sanity check of the code that bridges our data to their implementation. Even with his help, we spotted no faults. We want to invite any reader to check the publicly available code (see Section \ref{sec:introduction}). Furthermore, we asked the author for hints to tune ENCO for better performance which we also incorporated, but that led to no significant change in performance.

We argue that these surprising results for the benchmark still do not undermine the claims made in this paper. The improvement in performance w.r.t. the SOTA is not the core contribution of this paper and merely indicates that our novel approach performs empirically well. That being said, future research could look into the cause of these anomalies of ENCO when presented with the environments described in Section \ref{sec:MCD}.
\end{document}